\documentclass[a4paper]{article}

\usepackage[english]{babel}
\usepackage[utf8x]{inputenc}
\usepackage[T1]{fontenc}

\usepackage[a4paper,top=3cm,bottom=2cm,left=3cm,right=3cm,marginparwidth=1.75cm]{geometry}

\usepackage{amsmath}
\usepackage{graphicx}
\usepackage[colorinlistoftodos]{todonotes}
\usepackage[colorlinks=true, allcolors=blue]{hyperref}
\usepackage{relsize}
\usepackage{url}

\usepackage{subfig}
\usepackage{xcolor}     
\usepackage{floatrow}

\newcommand{\etal}{et~\textit{al}.}

\title{{Sit-to-Stand Analysis in the Wild using Silhouettes for Longitudinal Health Monitoring}}

\author{
  Alessandro Masullo\footnote{a.masullo@bristol.ac.uk}
  \and
  Tilo Burghardt
  \and
  Toby Perrett
  \and
  Dima Damen
  \and
  Majid Mirmehdi
}

\begin{document}
\maketitle

\begin{abstract}
We present the first fully automated Sit-to-Stand or Stand-to-Sit (StS) analysis framework for long-term monitoring of patients in free-living environments using video silhouettes. Our method adopts a {coarse-to-fine} time localisation approach, where a deep learning classifier identifies possible StS sequences from silhouettes, and a smart peak detection stage  provides fine localisation based on 3D bounding boxes. We tested our method on data from real homes of participants and monitored patients undergoing total hip or knee replacement. Our results show 94.4\% overall accuracy in the coarse localisation and an error of 0.026 m/s in the speed of ascent measurement, highlighting important trends in the  {recuperation} of patients who underwent surgery.
\end{abstract}

\section{Introduction}
Novel concepts and technologies like the Internet of Things (IoT) for Ambient Assisted Living (AAL) or specific health monitoring enable people to live independently, {to be aided in their recuperation, and} improve their quality of life. Such systems often include multiple sensors and monitoring devices, producing large amounts of data that need to be analysed and summarised in a few, clinically relevant parameters \cite{Zhu2015a}. The transition from a sitting position to a standing one (StS\footnote{In this work, {by StS we do in fact mean both `Sit-to-Stand' and `Stand-to-Sit', but will specify which of the two, if and when necessary.}}) is one of the most essential movements in daily {activities~\cite{Cheng2014}}, especially for older patients {suffering from musculoskeletal illnesses}. StS has been linked to recurrent falls \cite{Buatois2008}, sedentary behaviour \cite{Dall2010} and fall histories \cite{Yamada2013}. Continuous monitoring of the StS action over a long period of time can therefore highlight important trends, particularly for subjects undergoing physical rehabilitation.

{To the best of our knowledge, the automatic analysis of StS has not been attempted for long term monitoring and trend analysis. {Some previous works} have focused on automating the Sit-to-Stand clinical test, performed under supervised conditions and often in the presence of a clinician, e.g. \cite{Bohannon1995}.} {Shia~\etal{}~\cite{Shia2015}} suggested modelling the physics of {the} human body during  stand-up transitions by using a motion capture suite. Their method was tested in the lab on 10 healthy individuals but this approach is clearly impractical for long-term monitoring. In \cite{Galna2014}, Galna \etal{} investigated the suitability of skeleton data extracted by the Kinect sensor to assess clinically relevant movements, showing that the StS timing can be accurately captured with errors comparable to the VICON {motion capture} system. The{ir} method was applied in the lab on 9 individuals with Parkinson's Disease and 10 control subjects. Skeleton data was also used in \cite{ejupi2015kinect} to estimate the StS timing by using the vertical displacement of the head joint and a manual threshold. {The{ir} method was tested in the laboratory for 94~subjects and in participants' own homes for 20 individuals.} 

The {detection} of StS transitions can be seen as an action classification problem, and a large body of research has investigated the application of deep convolutional neural networks (CNN) for this task, for example \cite{Simonyan2014,Carreira2017}. However, while these works enable high accuracy in action classification, they always make use of RGB or depth data, which is not compatible with the privacy requirements of home monitoring systems, for {instance~}\cite{Ziefle2011,Birchley2017}. As already addressed in \cite{Calorinet2018}, silhouettes constitute a valid alternative form of data that allows {action recognition to be performed whilst}  respecting privacy requirements.

The aim of this work is to propose a novel approach to continuously monitor StS transitions in the wild {and, while addressing privacy issues,} to generate automatic trend analysis. {For each StS transition, we measure the speed of {ascent/descent} as an indicator of physical function.} We installed RGBD cameras~{(PrimeSense) in participants' own houses and} recorded silhouette video data from 9 subjects in 4 different habitations, for a minimum period of 4 months, up to 1 year, under the auspices of the SPHERE and HemiSPHERE projects~\cite{Grant2018,Zhu2015a}. Two of the participants, {aged between 65 and 90,} underwent total hip or knee replacement and we monitored them before and after their intervention. {The remaining 7 participants, aged between 40 and 60, did not record any particular health condition that could affect their mobility. }
We show that our method can identify StS transitions into the wild with 94.4\% overall accuracy and our measurement of the speed of ascent is comparable with the VICON motion capture gold standard in a supervised setting. 
Moreover, our analyses highlight important trends linked to the rehabilitation process, potentially allowing for surgeons to follow the progress of their patients remotely and anticipate possible complications.

\section{Methodology}
\label{sec:methodology}
Monitoring people in their homes poses {stringent} ethical restrictions on the type of data that can be recorded, analysed and shared, e.g. prohibiting the use of RGB data \cite{Sanchez2017,Zagler2008}. To provide a privacy-compatible monitoring system (based on a user study \cite{Zhu2015a}), we generate silhouettes and 3D bounding boxes from the RGB data and discard the raw pixel values immediately thereafter. 
{We deployed one camera in each house (in the living room) 
and set it up at a similar height to have a comparable field of view.}

\begin{figure}
    \centering
    \includegraphics[width=\textwidth]{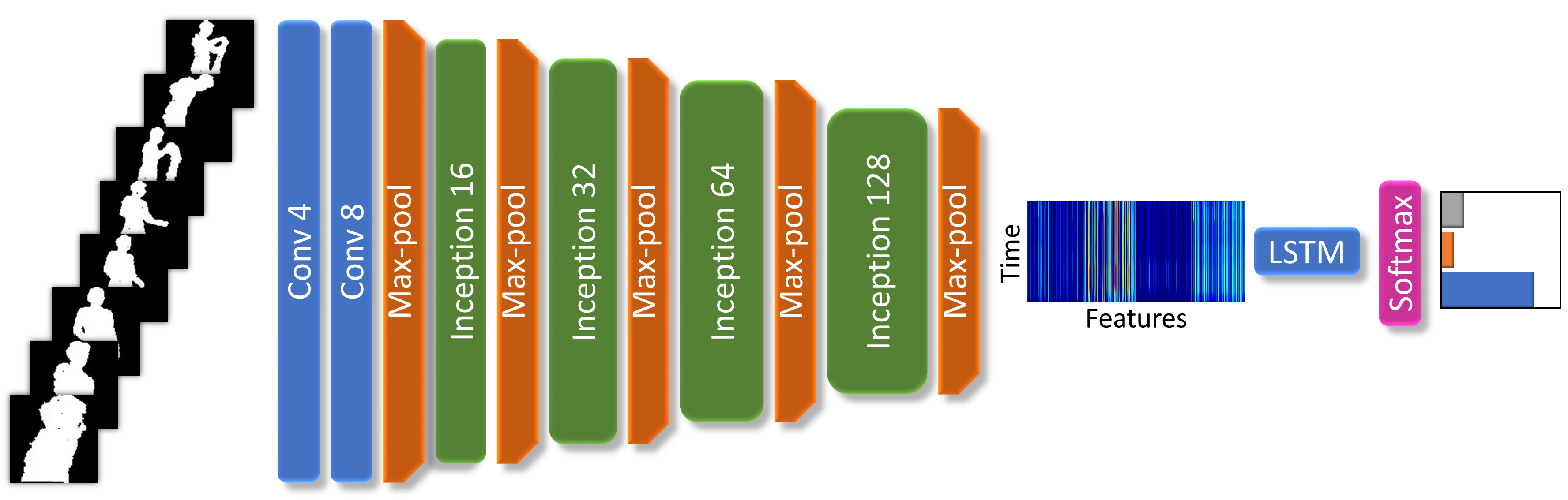}
    \caption{{Network architecture of the proposed method.}}
    \label{fig:network}
\end{figure}

{Our proposed pipeline can be divided into three steps: pre-processing of videos, classification and StS measurement. }First, the incoming silhouettes are cropped at the detected bounding boxes and resized, producing one video per individual. These videos are subdivided into short clips of 10 seconds each\footnote{{The frame-rate of the silhouette recorder varied according to different conditions and produced 10 fps on average}}, which are then classified with a deep CNN (detailed in Section~\ref{sec:classification}) into one of three categories: ``Sit-to-Stand'', ``Stand-to-Sit'' or ``Other''. The StS video clips only are then further analysed to measure the speed of {ascent/descent} using the 3D bounding boxes, as described in Section~\ref{sec:speed_of_ascent_measurement}. 

Contrary to previous works that have focused on StS duration \cite{Bohannon1995,Shia2015}, 
our method measures the speed of {ascent/descent}, defined as the maximal transferring velocity of the centre of gravity {(CG)} between the start and the completion of the StS movement \cite{Schot2003}.  
The speed of {ascent/descent} does not depend on a specific beginning or end of the movement, but rather on the maximum velocity. Thanks to this property, the speed of {ascent/descent} shows no significant difference between the Sit-to-Stand and the Sit-to-Walk movements \cite{Kouta2006}, {or the Stand-to-Sit and the Walk-to-Sit movements}, making it a more suitable measurement for free-living monitoring.

\subsection{Classification}
\label{sec:classification}

Inspired by the work from Carreira \etal{} \cite{Carreira2017}, we built our classifier network using Inception  modules with 3D convolutions, {as presented in Figure~\ref{fig:network}}. 
It was shown in our previous work \cite{Calorinet2018} that using very deep networks on silhouette data increases the computational cost without inducing any advantages. 
We therefore adopted a shallow architecture composed of 4 stacks of Inception modules, followed by a Long Short-Term Memory (LSTM) layer located between the last convolutional layer and the final fully connected layer. In our experiments, we found that the use of an LSTM module in addition to the 3D convolution produced the best results in classification accuracy.

The video sequences recorded from the participants' homes contained highly varied data, with video clips of StS transitions only constituting less than 1\% of the whole dataset. To tackle this class imbalance problem  \cite{Japkowicz2002}, we under-sampled the ``Other'' class to match the size of the minority classes ``Sit-to-Stand'' and ``Stand-to-Sit'', sampling new random elements for each epoch. This ensured a balanced training and prevented the potential loss of useful data from the ``Other'' class.

\subsection{Speed of Ascent Measurement}
\label{sec:speed_of_ascent_measurement}
\begin{figure}[t!]
    \centering
    \includegraphics[width=\textwidth,trim={0.4cm, 0.4cm, 0.4cm, 0.5cm},clip]{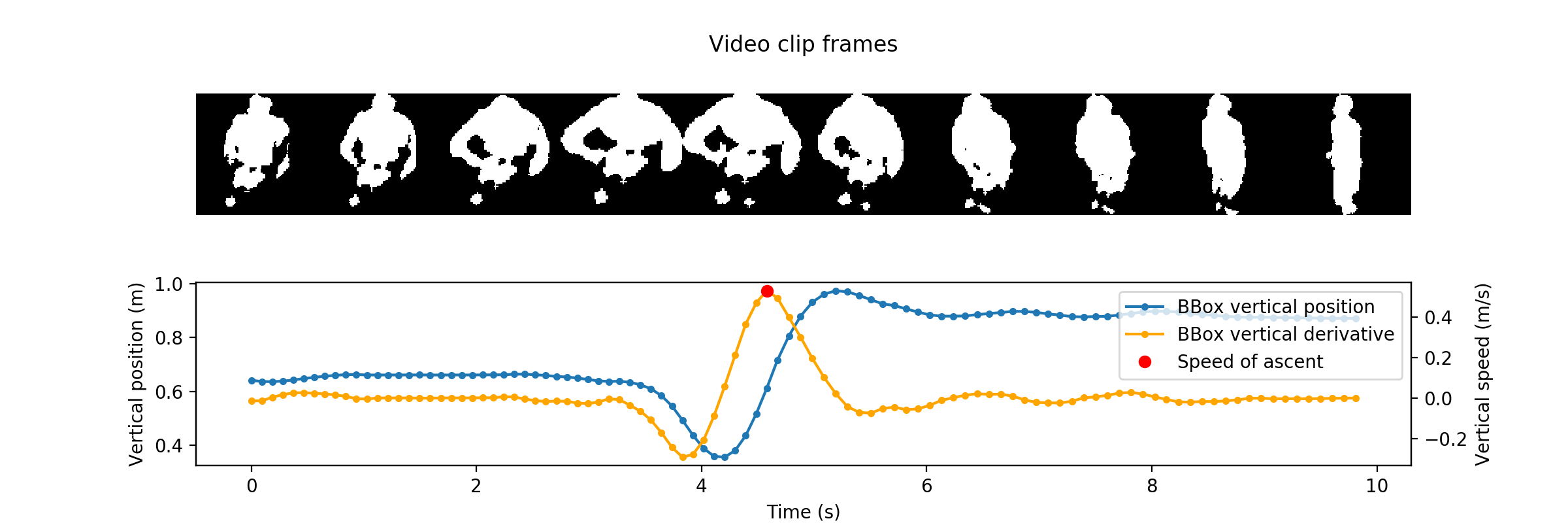}
    \caption{Example computation of the speed of ascent: (top) video frames {of a Sit-to-Stand sequence from the SPHERE data, }colour coded with intensity of the vertical derivative; (bottom) 3D bounding box vertical coordinate and derivative. {The maximum intensity of the vertical speed corresponds to the speed of ascent.}}
    \label{fig:vertical_derivative}
\end{figure}

The 10-seconds clip classifier provides a coarse time localisation of the StS transitions. To narrow the exact frame of the transition and measure the speed of ascent\footnote{{Although here we refer to the computation of the speed of ascent, the methodology applies identically for the speed of descent by simply using the negative sign in Eq.~\ref{eq:derivative}.}}, we employ data from the 3D bounding boxes, in particular the evolution in time of the upper edge. 
Let the 3D bounding box $B$ for the time interval $[t_\text{start},\ t_\text{end}]$ of a clip be 
$B\left(t\right) = \left[x_1(t), y_1(t), z_1(t); x_2(t), y_2(t), z_2(t)\right]$, where the indices 1 and 2 respectively represent the `right', `top', `front', and `left', `bottom', `back' vertices of the 3D box. Let us call $y_1 \equiv y_\text{top}$ the $y$ component of the top vertex, and the vertical speed of the subject can then be estimated as:
\begin{equation}
\label{eq:derivative}
    V_y(t) = \pm\frac{d y_\text{top}}{dt} ~,
\end{equation}
where the sign is $+$ for ``Sit-to-Stand'' and $-$ for ``Stand-to-Sit'' classes. Using the definition of speed of ascent as the maximum vertical velocity during the StS movement, we can then compute the speed of ascent $V_{\text{SOA}}$ as: 
\begin{equation}
\label{eq:speed_of_ascent}
    V_\text{SOA} = \max_{[t_\text{start},\ t_\text{end}]}\left\{V_y(t)\right\}.
\end{equation}
It is important to note that the computation of Eq. (\ref{eq:speed_of_ascent}) is only performed on those clips classified earlier as StS. In fact, its simplicity is {built upon} 
 the accuracy of the classifier,  which filters out all the other possible movements that might contain a vertical motion and are not StS transitions. A visualisation of this computation can be seen in Figure~\ref{fig:vertical_derivative}, showing a strong correlation between the vertical speed of the bounding box and the Sit-to-Stand action.

In order to reduce noise of the 3D bounding boxes, we adopted a Savitzky-Golay filter (\verb|savgol|) as implemented in SciPy. The advantage of the \verb|savgol| filter is that it replaces each data-point by the least-squares polynomial fit of its neighbours, allowing noise reduction and a simple analytical derivative of the polynomial. We used a kernel window size of 11 points and a polynomial of 3rd order. The vertical velocity can then be computed as the ratio between the filtered $y_\text{top}$ and the filtered time vector:
\begin{equation}
    \label{eq:savgol_deriv}
    V_y = \frac{\text{savgol}(y_\text{top},\text{ deriv=1})}{\text{savgol}(t,\text{ deriv=1})}
\end{equation}

\section{Experiments}

The architecture was built with 4 Inception modules \cite{Szegedy2015}, each composed of a sequence of (1) 3D convolutions, (2) batch normalisation and (3) activation ReLu, using respectively 16, 32, 64 and 128 filters. The last layer produces a set of convolutional features which are, once reshaped, 512 dimensional for 25 pseudo-time steps. The resulting features are fed into an LSTM module with 128 units, whose output is then fed into a 3D fully connected layer with \textit{softmax} activation.
The input comprises video clips of 100 frames, each 100 by 100 pixels, while the output is a 3 by 1 classifier.

We demonstrate the validity of our algorithm by assessing the StS video classifier and the speed of {ascent/descent} computation independently on two different datasets.

\subsection{Physical Rehabilitation Movements Data Set}
\label{sec:validation_UIPRMD}
The UI-PRMD dataset includes skeleton data from typical exercises and movements which are performed by patients during therapy and rehabilitation programs \cite{Vakanski2018}. It consists of 10 healthy subjects, performing 10 different movements 10 times each, and recorded simultaneously using a Kinect and a VICON (gold standard) motion-capture system.

In particular for our work, we extracted the Sit-to-Stand movement from the dataset and used the VICON motion capture data to validate our proposed approach. 
{We generated 3D bounding boxes using the extent of Kinect skeleton joints and we compared the speed of ascent with the one computed using the centre of gravity (CG) from the VICON data}\footnote{The CG was estimated using the average of the Left, Right, Anterior and Posterior Superior Illiac skeletal joints.}.

\begin{figure}
    \centering
    \subfloat[]{
        \includegraphics[width=0.6\textwidth]{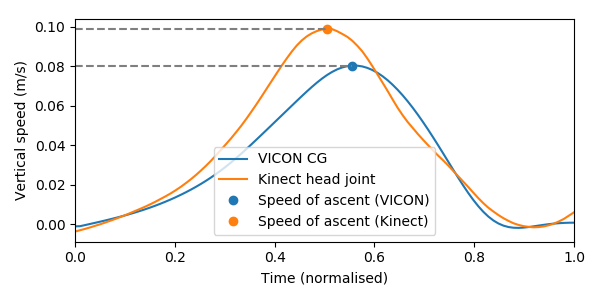}
    }
    \subfloat[]{
        \includegraphics[width=0.4\textwidth]{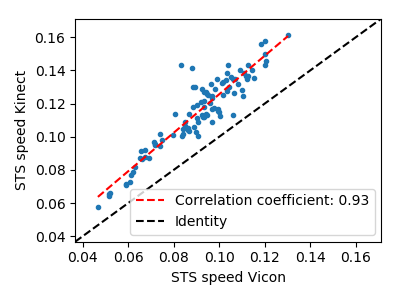}
    }
    \caption{Comparison of speed of ascent computed with our algorithm using the Kinect data and the VICON system}
    \label{fig:kinect_vs_VICON}
\end{figure}

{The curves in Figure~\ref{fig:kinect_vs_VICON}a show a comparison of the true speed of ascent, computed using the VICON CG (blue curve), and our estimation using the Kinect head joint (orange). In both cases, the vertical derivative was obtained for all the StS transitions available ($N_\text{StS}=100$) and averaged to highlight possible discrepancies, while the time was normalised using the beginning and the end of the StS trasition.} 
The two curves exhibit a very similar pattern, with a maximum value {(i.e. the speed of ascent)} which differs by 23.3\%. 
This amplification of the maximum vertical speed results in a bias error of the speed of ascent of about 0.026 m/s, or 28.3\% of the average measurement. In spite of this bias error, the correlation between our estimated speed of ascent and the ground truth is more than 92.8\%, as shown in Figure~\ref{fig:kinect_vs_VICON}b. While this bias could be mitigated by appropriate calibration, the aim of this work is to investigate trends in the speed of {ascent/descent} and the high correlation between our measurement and the ground truth is more than sufficient for its application.

\subsection{SPHERE data}
The SPHERE project (Sensor Platform for Healthcare in a Residential Environment) \cite{Zhu2015a} developed a multi-modal sensing platform aimed to record data from up to 100 houses in the Bristol (UK) area for healthcare monitoring. Each house was equipped with a variety of sensors, including RGBD cameras, which were used to generate human {silhouettes and 2D/3D bounding boxes via the OpenNI API \cite{OpenNI}}, from different communal spaces: living room, kitchen and hall. The HEmiSPHERE (Hip and knEe study of SPHERE) project \cite{Grant2018} is an {UK National Health Service} application of SPHERE sensors within the homes of patients undergoing a total hip or knee replacement.

In this work, we present data collected from the living room of 4 different houses, described in Table~\ref{tab:houses}, two belonging to the HEmiSPHERE cohort and two belonging to the SPHERE one. This subset includes a total of 1,177,082 video clips, of which 5,645 are StS transitions and the rest belong to the ``Other'' class. 
The videos were manually labelled by the authors using the MuViLab annotator tool\footnote{Available on GitHub: \url{https://github.com/ale152/muvilab}} and were used for cross-validation as per Table~\ref{tab:results}.
The discrepancy between the number of Sit-to-Stand and Stand-to-Sit transitions can be explained by the type of silhouette detector adopted (OpenNI), that was optimised for standing poses. This increases the chances of detecting a person walking and sitting down and hence the number of Stand-to-Sit transitions recorded. 

\begin{table}[]
\centering
\begin{tabular}{l|c|c|c|c|c}
\hline
\multicolumn{1}{c|}{\bf Id}  & \multicolumn{1}{c|}{\bf Duration} & {\bf Occup.} & {\bf \#Other} & {\bf \ \#Sit-to-Stand} & \multicolumn{1}{c}{\bf \#Stand-to-Sit} \\ \hline
{\bf House {\it A}}    & 4 months             & 2      & 107404  & 339     & 491                       \\
{\bf House {\it B}}    & 3 months             & 2      & 266853  & 1289    & 2051                      \\
{\bf House {\it  C}}    & 9 months             & 4      & 416628  & 297     & 1054                      \\
{\bf House {\it D}}    & 6 months             & 1      & 380552  & 54      & 70                        \\ \hline
\end{tabular}
\caption{Description of the data from the 4 houses: 2 cohorts of SPHERE (bottom two rows) and HemiSPHERE (top two rows).}
\label{tab:houses}
\end{table}

\begin{figure}
\centering
\subfloat[Fold 1]{
    \includegraphics[width=0.33\textwidth,trim={0.1cm, 0.5cm, 0.3cm, 0.5cm},clip]{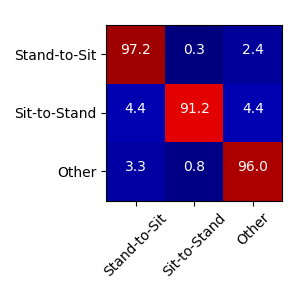}
}
\subfloat[Fold 2]
{
    \includegraphics[width=0.33\textwidth,trim={0.1cm, 0.5cm, 0.3cm, 0.5cm},clip]{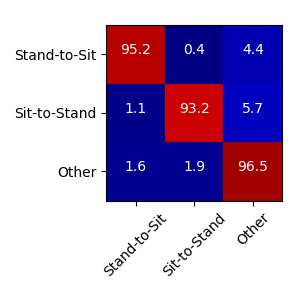}
}
\subfloat[Fold 3]
{
    \includegraphics[width=0.33\textwidth,trim={0.1cm, 0.5cm, 0.3cm, 0.5cm},clip]{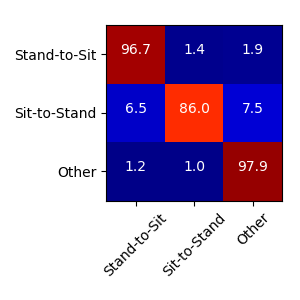}
}
\caption{Confusion matrix for each validation fold.}
\label{fig:confusion_matrices}
\end{figure}

\subsection{Classification}
Data from homes \textit{A}, \textit{B} and \textit{C} was used to train {and validate} the network (described in Section~\ref{sec:classification}) using a cross-validation strategy, as depicted in Table~\ref{tab:results}. {Data from \textit{House D} was left out of this procedure and was only used to generate the trend plot}. Results are presented in Table~\ref{tab:results} and show {an overall accuracy of 94.8\%, 95.0\% and 93.5\% for the three validation folds, computed by averaging the accuracy of the three classes. The average accuracy across the three folds is 94.4\%.} 
Details of the {classification results} are presented in Figure~\ref{fig:confusion_matrices}, showing the confusion matrices for each validation fold.

Particular attention must be paid to the false positive scores. The number of ``Other'' videos mis-classified as StS was found to be {1.63\%}, producing {28119} false positive against the {6548} correctly identified StS transitions. While these values might potentially damage our score, a manual inspection of the false positives concluded that many of the mis-classification videos are, indeed, visually similar to StS transitions. This included subjects interacting with the environment for long periods of time while standing up, raising from the floor, kneeling while doing exercises or housekeeping chores. Although these movements are not strictly StS transitions, they still involve a vertical motion that requires physical effort. As we will show in the next Section, although the presence of these false detection increases the uncertainty of our measurements, it does not hamper the calculation of the trend plots.

\begin{table}[]
\centering
\begin{tabular}{c|c|c|c|c|c|c}
\hline
{\bf Fold} & {\bf Train}     & {\bf Validate} & {\bf Stand-to-Sit}  & {\bf Sit-to-Stand}  & {\bf Other} & {\bf Overall} \\ \hline
1    & House C, B & House A   & 97.2\%         & 91.2\%         & 96.0\%      & 94.8\%    \\
2    & House C, A & House B   & 95.2\%         & 93.2\%         & 96.5\%      & 95.0\%    \\
3    & House A, B & House C   & 96.7\%         & 86.0\%         & 97.9\%      & 93.5\%    \\ \hline
\multicolumn{6}{r|}{Average} & 94.4\%  \\  
\end{tabular}
\caption{Cross-validation accuracy results, with $94.4\%$ overall average accuracy.}
\label{tab:results}
\end{table}

\subsection{Trend plots}
\label{sec:result_trend_plots}
Following the classification, the speed of {ascent/descent} was computed for all the video clips detected as StS transitions and it was averaged per week. The resulting trend plot, for \textit{Fold 2} as an example, is presented for the manually labelled video clips (\textit{Manual trend}) in Figure~\ref{fig:trend_comparison_hemisphere}a, and for the automatic labels (\textit{Automatic trend}) in Figure~\ref{fig:trend_comparison_hemisphere}b. The reader is reminded that one of the occupiers of this house underwent a total hip or knee replacement intervention and the surgery day is marked with a solid black line. Before surgery, the speed of ascent is between 0.35 and 0.45 m/s, which is followed by a sudden drop soon after the operation. This is due to the pain and the discomfort following the surgery, which impair the physical ability of the patient  and hence their speed of ascent. In the following weeks, the speed of ascent shows a slow but steady increase with a slope of around 0.04 m/s per month. Finally, 14 weeks after the surgery, the speed of ascent reaches a value which is just shy of 0.5 m/s, confirming a full recovery. {The presence of the trend is also corroborated by a high coefficient of determination $R^2=0.86$.}

The comparison between the \textit{Manual trend} and the \textit{Automatic trend} from Figure~\ref{fig:trend_comparison_hemisphere} shows a very similar pattern, {with a correlation coefficient between the two plots of 0.88}. In spite of the higher error bars, due to false positives, the main characteristic aspects of the plot are preserved, including the drop in the speed of ascent following the surgery and the full recovery after 14 weeks. 

\newcommand{\myysize}{0.45}
\begin{figure}
    \centering
    \subfloat[Manual]{
    	\includegraphics[width=\myysize\textwidth]{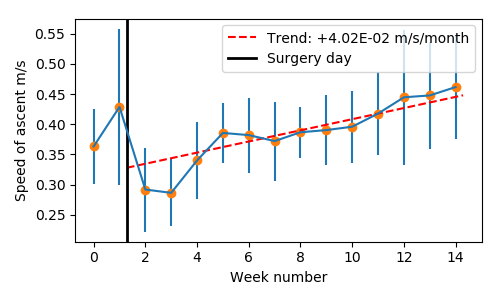}
    }
    \subfloat[Automatic]{
    	\includegraphics[width=\myysize\textwidth]{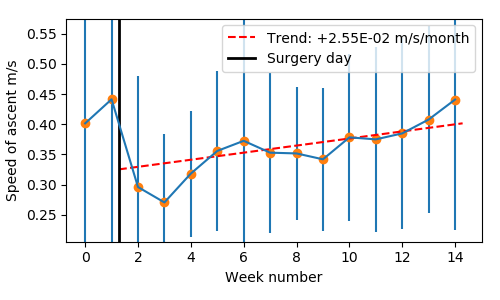}
    }
    \caption{Comparison of speed of ascent trend for Fold 2, extracted from (a) the manually labelled StS transitions and (b) the video clips automatically labelled as StS. {The correlation between the plots is 0.88.}}
    \label{fig:trend_comparison_hemisphere}
\end{figure}

For comparison, we present \textit{Automatic trends} generated for \textit{House C} and \textit{D} in Figure~\ref{fig:trend_comparison_sphere}, occupied by healthy participants. As expected, no particular trend can be noticed for these houses, {as confirmed by the low coefficients of determination $R^2$  of -0.21 and -0.45 {respectively}.}

\newcommand{\myysizee}{0.45} 
\begin{figure}
    \centering
    \subfloat[House C]{
    	\includegraphics[width=\myysizee\textwidth]{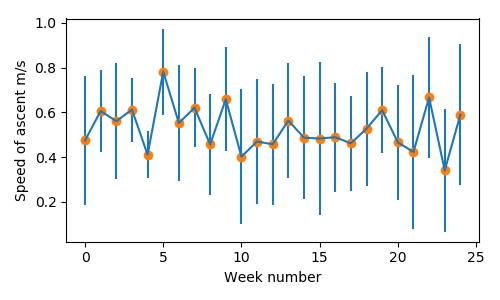}
    }
    \subfloat[House D]{
    	\includegraphics[width=\myysizee\textwidth]{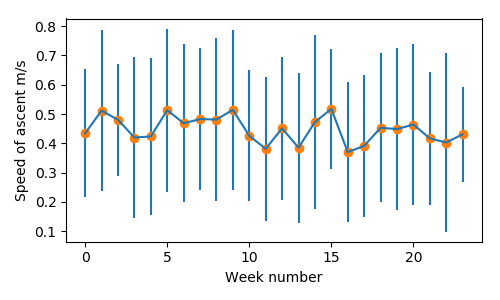}
    }
    \caption{Comparison of speed of ascent trend for \textit{House C} and \textit{D} from the SPHERE cohort.}
    \label{fig:trend_comparison_sphere}
\end{figure}

{Although the trend plots presented in this section only refer to the speed of ascent (i.e. Sit-to-Stand), the trend plot computed using the speed of descent (i.e. Stand-to-Sit) showed a very similar behaviour and were omitted from this paper for brevity.}

\section{Conclusions}
The demand of AAL technologies for home monitoring is continuously increasing. We presented a simple and efficient approach for the detection and analysis of StS transitions for home monitoring in completely unsupervised environments. We implemented and tested our method in 4 different houses, 2 of which were occupied by patients with total hip or knee replacement. We showed that we are able to reliably identify StS transitions in video clips of binary silhouettes and we can confidently measure the speed of ascent for each transition as an indicator of improving or deteriorating functionality for the StS test. Plots of the average speed of ascent estimated by our method highlights important trends in the recovery process of the surgery patients.

\section{Acknowledgements}
This work was performed under the SPHERE IRC funded by the UK Engineering and Physical Sciences Research Council (EPSRC), Grant EP/K031910/1. The authors wish to thank {all the study subjects for their participation in this project and} Rachael Gooberman-Hill, Andrew Judge, Ian Craddock, Ashley Blom, Michael Whitehouse and Sabrina Grant for their support with the HEmiSPHERE project. The HEmiSPHERE project was approved by the Research Ethics Committee (reference number: 17/SW/0121).
%
%

\bibliographystyle{plain}





\end{document}